\documentclass[conference]{IEEEtran}

\let\labelindent\relax
\IEEEoverridecommandlockouts 
\usepackage{times}
\usepackage{mwe}
\usepackage[numbers]{natbib}
\usepackage{multicol}
\usepackage[bookmarks=true]{hyperref}
\usepackage{color}
\usepackage{enumitem}
\usepackage{graphicx}
\usepackage{amsmath}
\usepackage{balance}
\usepackage{
tikz,
relsize,
booktabs
}
\usepackage{amssymb}
\usepackage{algorithmic}
\usepackage[font={small}]{caption}
\usepackage{xcolor}
\usepackage{hyperref}
\usepackage{pifont}
\usepackage{color}
\usepackage{colortbl}
\usepackage[table]{xcolor} 

\newcommand{\method}{\textsc{Aina}}
\newcommand{\website}{\url{https://aina-robot.github.io}}

\title{\LARGE \bf
Dexterity from Smart Lenses: Multi-Fingered Robot Manipulation with In-the-Wild Human Demonstrations
}

\author{
Irmak Guzey$^{1,2}$ \qquad Haozhi Qi$^{2}$  \qquad Julen Urain$^{2}$ \qquad Changhao Wang$^{2}$ \qquad Jessica Yin$^{2}$ \\
\qquad Krishna Bodduluri$^{2}$ \qquad Mike Lambeta$^{2}$ \qquad Lerrel Pinto$^{1}$  \qquad Akshara Rai$^{2}$ \\ \qquad Jitendra Malik$^{2}$ \qquad Tingfan Wu$^{2}$ \qquad Akash Sharma$^{2}$ \qquad Homanga Bharadhwaj$^{2}$
\\ \\ $^{1}$ New York University, $^{2}$ Meta
\\ \\ { \tt \href{https://aina-robot.github.io}{aina-robot.github.io}}\\
\thanks{Correspondence to \texttt{irmakguzey@nyu.edu}.}
}

\definecolor{blue}{HTML}{0173B2}
\definecolor{green}{HTML}{06A66C}
\definecolor{red}{HTML}{CF3C33}
\definecolor{magenta}{HTML}{D748A2}
\hypersetup{
    colorlinks=true,
    linkcolor=blue,
    filecolor=magenta,      
    citecolor=blue,
    urlcolor=teal,
}

\begin{document}

\newcommand{\xxnote}[3]{}
\ifx\hidenotes\undefined
  \renewcommand{\xxnote}[3]{\color{#2}{#1: #3}}
\fi

\newcommand{\IG}[1]{{\xxnote{IG}{magenta}{#1}}}

\makeatletter
\let\@oldmaketitle\@maketitle%
\renewcommand{\@maketitle}{\@oldmaketitle%
    \centering
    \includegraphics[width=\linewidth]{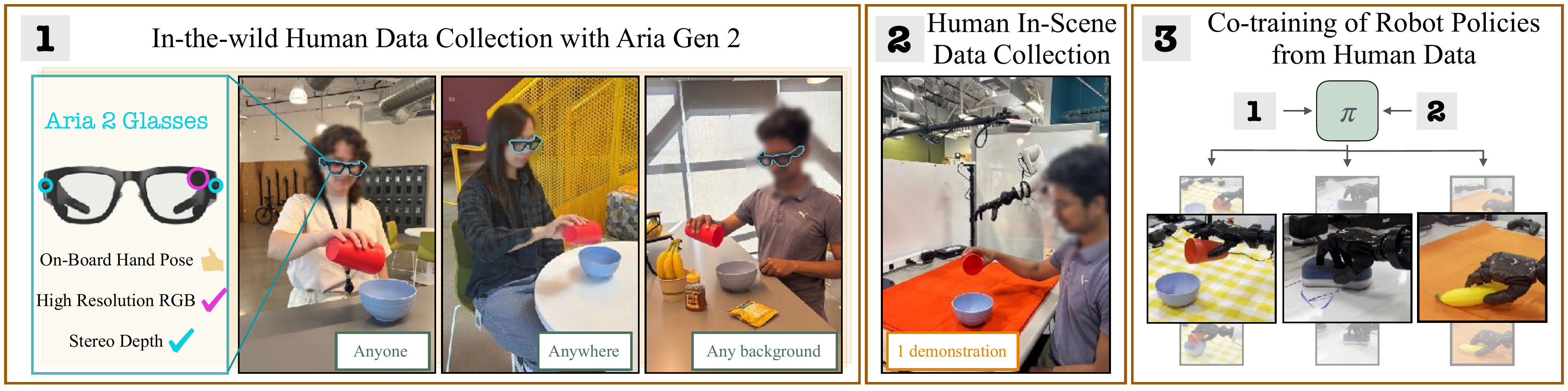}
    \captionof{figure}{\method{} is a framework for learning multi-fingered policies from in-the-wild human data collected with smart glasses, without requiring any robot data (including online corrections or simulation). The workflow is as follows: a human wears the Aria 2 glasses and collects in-the-wild demonstrations on any surface with arbitrary backgrounds (left), then records a single demonstration in the robot deployment space (middle), after which point-based policies are trained and directly deployed on the robot (right). With an average of just 15 minutes of human video collection effort, \method{} is able to train autonomous robot policies.}
    \label{fig:intro}
}

\maketitle
\begin{abstract}
Learning multi-fingered robot policies from humans performing daily tasks in natural environments has long been a grand goal in the robotics community. Achieving this would mark significant progress toward generalizable robot manipulation in human environments, as it would reduce the reliance on labor-intensive robot data collection. Despite substantial efforts, progress toward this goal has been bottle-necked by the embodiment gap between humans and robots, as well as by difficulties in extracting relevant contextual and motion cues that enable learning of autonomous policies from in-the-wild human videos. We claim that with simple yet sufficiently powerful hardware for obtaining human data and our proposed framework \method{}, we are now one significant step closer to achieving this dream. \method{} enables learning multi-fingered policies from data collected by anyone, anywhere, and in any environment using Aria Gen 2 glasses. These glasses are lightweight and portable, feature a high-resolution RGB camera, provide accurate on-board 3D head and hand poses, and offer a wide stereo view that can be leveraged for depth estimation of the scene. This setup enables the learning of 3D point-based policies for multi-fingered hands that are robust to background changes and can be deployed directly without requiring any robot data (including online corrections, reinforcement learning, or simulation). We compare our framework against prior human-to-robot policy learning approaches, ablate our design choices, and demonstrate results across nine everyday manipulation tasks. Robot rollouts are best viewed on our website: \website{}. 
\end{abstract}

\IEEEpeerreviewmaketitle

\section{Introduction}

\begin{figure*}[t]
    \centering
    \includegraphics[width=\linewidth]{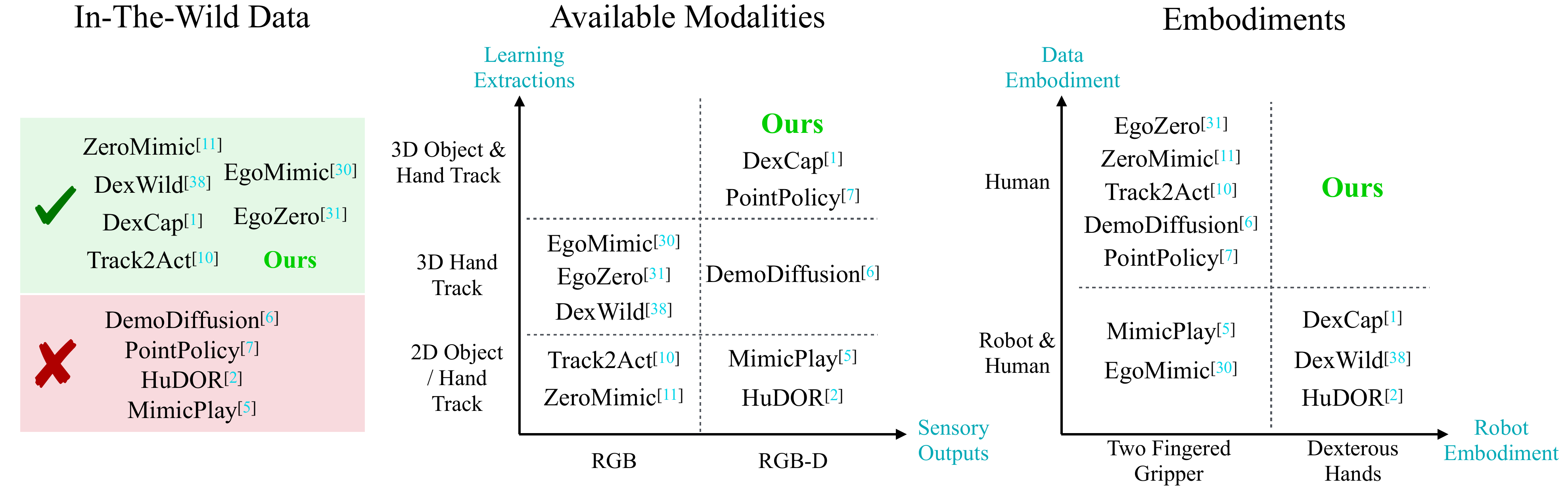}
    \caption{Comparison of \method{}'s capabilities with some prior human-to-robot learning frameworks. \textit{In-The-Wild} indicates whether data can be easily collected in natural settings outside the lab. \textit{Sensors} describes the sensory outputs available from the data collection devices. \textit{Learning Extractions} specifies which extractions can be utilized with the provided sensors to improve learning. \textit{Data Embodiment} refers to the embodiment of the collected data (robot vs. human). Here, we also count online corrections~\cite{wang2024dexcap} and reinforcement learning~\cite{hudor} performed on the robot as part of the robot data. \textit{Robot Embodiment} indicates which type of robot embodiment the framework targets (two-fingered gripper vs. multi-fingered hand). In \method{}, we choose point-based approaches for their robustness to background variations, enabling robot learning from in-the-wild data for dexterous hands. This is made possible by the advanced sensing capabilities of the Aria Gen 2 glasses, which provide all the necessary 3D extractions.}

    \label{fig:comparison}
    \vspace{-15pt}
\end{figure*}

\begin{center}
\begin{minipage}{0.92\linewidth}
\begin{quote}\itshape
“The most profound technologies are those that disappear. They weave themselves into the fabric of everyday life until they are indistinguishable from it.”
\par\raggedleft — Mark Weiser, 1991
\end{quote}
\end{minipage}
\end{center}

Robots autonomously performing diverse manipulation tasks by watching humans go about their daily lives has been a dream in Artificial Intelligence (AI) for decades. However, this remains challenging due to the embodiment gap between humans and robots, as well as the disparity between human video views and the sensor perspectives of a robot. To truly realize this dream for generalizable dexterous manipulation, we must overcome these challenges with general approaches that can leverage large-scale human video data. Encouragingly, this vision is now closer to reality with the development of increasingly more human-like robot embodiments~\cite{ability-hand,zorin2025ruka} and the potential widespread adoption of wearable devices such as smart glasses, which are lightweight, easy to wear in daily life and equipped with complex sensing capabilities that provide both an egocentric perspective and rich annotations. Building on this promise, we develop an approach to learn dexterous manipulation directly from smart-glass human data, without requiring any additional robot interaction data.

We are, of course, not the first to consider this setting of learning manipulation from human videos. Prior work has attempted to address these challenges by collecting human videos in structured settings, often within the exact scenarios of robot deployment~\cite{wang2023mimicplay, demo-diffusion, point-policy}. However, such approaches are difficult to scale to diverse environments, as they require data collection for each deployment scenario. Other efforts leverage large-scale, in-the-wild web videos~\cite{bharadhwaj2023generalizable, vrb, track2act, zeromimic}, but they have not been successfully deployed on multi-fingered hands, since extracting the necessary annotations—such as reliable 3D hand poses—for learning dexterous policies is far more challenging in these settings. Smart-glasses data, offers the best of both worlds: it preserves scalability by being naturally collected as people go about daily life, while providing high-resolution egocentric imagery, stereo vision for 3D perception, and reliable hand-pose annotations via in-built software~\cite{nimble}. These characteristics make smart-glass data far richer and more robot manipulation-relevant than web video, while avoiding the scalability bottlenecks of lab-constrained data collection.

Thus, leveraging the complex sensing capabilities of smart glasses, in particular of Aria Gen 2, we develop \textbf{\method}: a simple approach for learning a closed-loop dexterous manipulation policy from just human videos. \method{} (english. Mirror) refers to mirroring human videos in a robot's context and is based on a simple intuition: By lifting human videos to approximate 4D via hand-keypoint reconstruction, stereo depth estimation, and 3D object pointcloud extraction, we can re-purpose 3D policy learning approaches for learning to predict future hand keypoints, and use the same policy for robot manipulation. By operating in the space of 3D keypoints for the hand, and 3D pointclouds for objects, we minimize the human-robot domain gap when deploying the \method~policy on a dexterous robot hand, while being trained with only human demonstrations.

Concretely, \method~operates as follows: (a) humans wearing smart glasses collect data in arbitrary environments with any background or viewpoint, (b) then they collect a single video demonstration in the robot’s environment, and (c) the multi-fingered robot learns policies that generalize across both spatial configurations and object variations. We evaluate \method{} on nine tasks and summarize our contributions as follows:
\begin{enumerate}
    \item \method{} is the first framework that learns policies for multi-fingered hands without using any robot data, including no use of simulation (Section~\ref{section:method}).
    \item \method{} leverages recent advances in computer vision techniques and smart glasses to accurately track hand and objects in 3D and learn closed-loop policies to transfer them to the robot environment. 
\end{enumerate}
In Section ~\ref{sec:experiments}, we show that \method{} outperforms existing human-to-robot learning approaches demonstrating the effectiveness of learning manipulation from human videos alone through a simple framework operating on rich sensing from smart glasses. Robot videos are available on our website: \website{}.
\begin{figure*}[t]
    \centering
    \includegraphics[width=\linewidth]{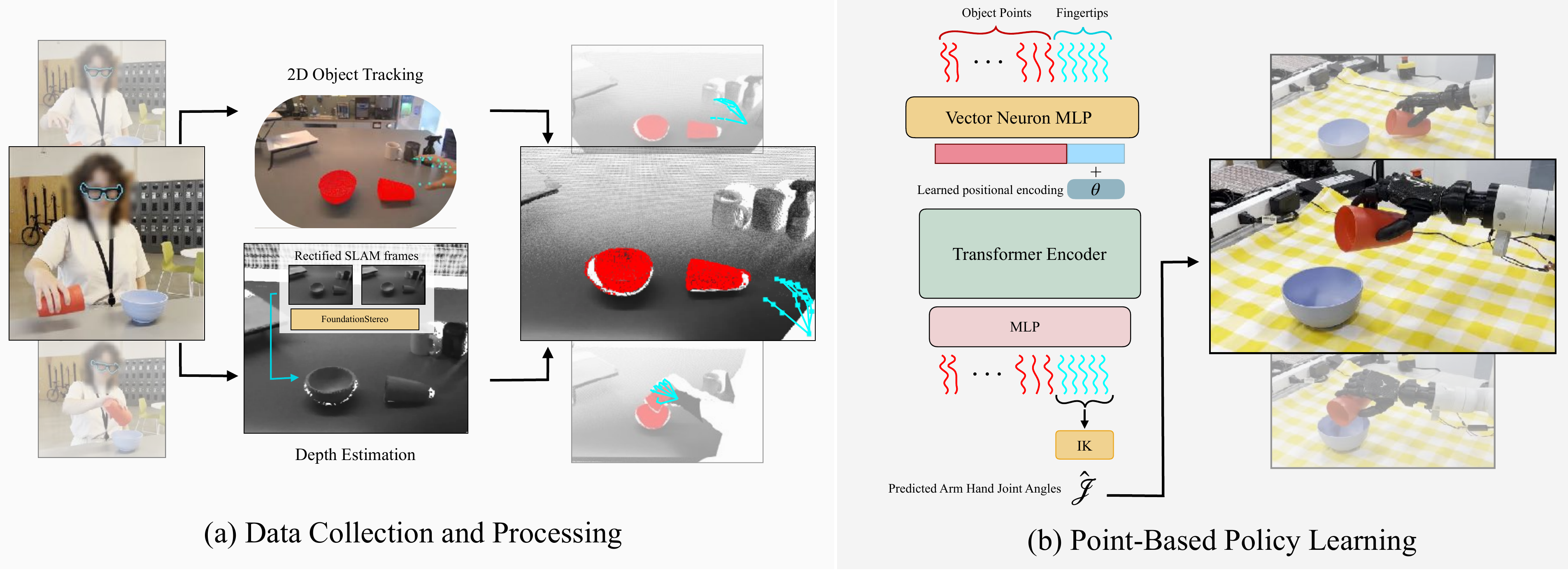}
    \caption{Illustration of our overall \method{} framework. On the left, we show how the data is processed: the human hand pose is extracted directly by the Aria Gen 2 glasses, and stereo depth is estimated from the surrounding SLAM camera frames. This enables the 3D policy learning methods on the right to succeed while remaining robust to background clutter.} 
    \label{fig:method}
\end{figure*}

\section{Related Works}
\label{sec:related}

\method{} draws inspiration from extensive research in dexterous manipulation, learning from human videos, and imitation learning. Our aim is to develop a \textit{simple} framework for closed-loop policy learning capable of performing diverse everyday manipulation tasks with dexterous multi-fingered hands. We highlight some of the comparisons with prior works in Fig.~\ref{fig:comparison} and describe them below.

\paragraph{Robot Learning with Non-Robot Datasets}

Since robot interaction data collection is challenging due to operational constraints~\cite{arunachalam2023holo, iyer2024open}, thanks to advances in representation learning~\cite{he2022masked, wu2023unleashing}, motion prediction~\cite{cotracker, tapir}, and hand–object reconstruction~\cite{ye2022s,pavlakos2024reconstructing}, many approaches now leverage non-robot datasets such as human videos and images.  These approaches differ both in the type of human data used—in-domain vs. in-the-wild—and in what is extracted or learned from such data.

In-domain demonstrations, collected in the same environment as deployment, allow rich extractions like 3D hand poses and object points~\cite{demo-diffusion, point-policy, hudor, wang2023mimicplay}, but require new data per deployment and are thus hard to scale. In contrast, in-the-wild human datasets~\cite{ego4d, banerjee2024hot3d, grauman2024ego} support broader generalization, with works focusing on visual backbones~\cite{r3m, pvr2, voltron} or high-level cues such as hand-object trajectories~\cite{track2act, shaw2023videodex, zeromimic} and affordances~\cite{srirama2024hrp, vrb, chen2025web2grasp}. Yet, without reliable low-level signals like 3D hand pose, these methods often sacrifice accuracy or need additional demonstrations during deployment~\cite{zeromimic}.

More recently, smart glasses~\cite{aria-gen-1} have simplified data collection~\cite{egomimic, egozero}, enabling richer extractions and better generalization which \method{} builds upon. However, most of these works focus on two-finger grippers, where manipulation can be modeled simpler. In \method{}, we use Aria Gen 2 glasses~\cite{aria-gen-2} for scalable human data collection, but uniquely demonstrate policy learning from purely human demonstrations for dexterous multi-fingered robot hands.

\paragraph{Dexterous Manipulation from Human Data}

Early research on dexterous manipulation relied on sim-to-real transfer~\cite{shaw2023leap, akkaya2019solving} or teleoperation for data collection~\cite{arunachalam2023holo, yang2024ace, iyer2024open, tavi}, but these approaches are limited either by sim-to-real gaps or by the extensive human effort required for teleoperation. To address these challenges and reduce dependence on large-scale robot demonstrations, recent work has shifted toward learning from human data. However, leveraging large-scale datasets is harder for multi-fingered hands due to the lack of annotations needed for extracting reliable signals. As a result, most prior works have collected their own human data to train policies. Some collected in-domain human videos and extracted 3D hand poses~\cite{hudor, chen2024arcap}, while others gathered in-the-wild demonstrations using portable custom hardware with multiple cameras and hand-pose estimators~\cite{wang2024dexcap, dexwild}. While promising, all of these approaches incorporated some robot data, obtained either through teleoperation~\cite{dexwild} or online corrections~\cite{hudor, wang2024dexcap}. Although such robot data can help in complex dexterous tasks—particularly given the absence of force feedback in human demonstrations—in \method{} we demonstrate how we can learn to perform everyday manipulation activities with dexterous multi-fingered hands with just offline human videos captured through Aria glasses, without using any external sensors, mocap markers, or exo-skeletons.

\paragraph{Policy Architectures for Imitation Learning} 
Going beyond the standard of 2D image-based policies~\cite{rt1, r3m, cacti} for imitation, recent works have proposed 3D policy architectures that exploit geometric structure for manipulation~\cite{shridhar2023perceiver, ze2023gnfactor, gervet2023act3d}, yielding improved generalization to cluttered scenes and complex object interactions. Beyond raw pixels and scene point clouds, some approaches incorporate intermediate object-centric representations such as keypoints or tracks. PointPolicy~\cite{point-policy} learns manipulation policies from 3D hand and object keypoints, while Track2Act~\cite{track2act} predicts future dense object tracks from video datasets and trains track-conditioned policies. These object-centric methods highlight the benefits of embodiment-agnostic cues for bridging human and robot domains. Building on this insight, our proposed approach,~\method{} extends 3D imitation learning frameworks by extracting hand keypoints and 3D object flow from human videos, enabling policies that generalize across embodiments and leverages (non-robot) human data for dexterous manipulation.
\section{Method}
\label{section:method}

\subsection{Overview}
\label{sec:method:overview}

\method{} is a framework for learning closed-loop policies from in-the-wild human demonstrations collected with Aria Gen 2 glasses, without requiring any robot data. Our framework consists of three high-level steps:
(a) a human collects in-the-wild video demonstrations on arbitrary surfaces using the Aria Gen 2 glasses, along with a single in-scene video in the robot’s environment;
(b) the dataset is processed to extract 3D object tracks and hand fingertip points, which are then aligned to establish a uniform reference frame with the robot; and
(c) point-based policies are trained and deployed on a single-arm hand robot system.
We describe the overall structure of our framework in Fig.~\ref{fig:method} and describe the assumptions, challenges, and details in this section. 

\paragraph{\textbf{Assumptions}} In \method{}, our methodology is guided by two key assumptions:
(a) access to a calibrated scene to ensure a uniform operational space. For this, we perform hand–eye calibration to compute the extrinsic matrix of cameras on the robot setup with respect to the robot base. This process is straightforward, performed only once, and takes approximately 5–10 minutes during the initial setup of the robot.
(b) access to a single in-scene demonstration along with multiple in-the-wild demonstrations. Both types of demonstration are collected by humans (without using the robot). The in-scene demonstration takes less than a minute to collect, while the in-the-wild demonstrations take about 10 minutes in total for 50 demos per task.

\paragraph{\textbf{Challenges}} Unlike prior works that artificially constrain hand motions to be robot-like~\cite{lepert2024shadow} or require additional alignment hardware such as ArUco markers~\cite{egozero}, our approach considers in-the-wild human videos as \textit{natural} interactions. Our method does not require prior knowledge of the distance to manipulated objects, and it places no restrictions on the hand motions of the data collectors~\cite{point-policy}. These relaxations introduce  challenges, as the hand motions are more varied and less structured.

\subsection{Collecting and Processing Smart Glass Data}

\subsubsection{Data Collection}
\label{sec:method:data_collection}
\method{} uses Project Aria Gen 2~\cite{aria-gen-2} glasses to collect in-the-wild human demonstrations. The glasses are equipped with a front-facing RGB camera, four SLAM cameras positioned around the frame, and multiple IMUs. These sensors enable real-time estimation of the user’s head pose as well as left and right hand poses~\cite{nimble}. The head pose is defined with respect to a world frame arbitrarily assigned at initialization. This world frame is initialized using the gravity vector measured by the IMUs~\cite{aria-gen-1,aria-gen-2}, ensuring that its $z$-axis is aligned with gravity. For each task, we collect 50 in-the-wild demonstrations using these glasses and record the camera streams along with head and hand pose estimates at 10 Hz.

During data collection, we do not assume any specific height for the surfaces where humans manipulate the objects. As a result, we need to ground in-the-wild demonstrations within the robot’s scene. To address this, we collect a single \textit{in-scene} demonstration using the RGB-D cameras in the robot environment. We estimate the hand pose in 2D from both camera views using Hamer~\cite{hamer,mano}, and then triangulate these estimates to obtain the 3D pose~\cite{point-policy, demo-diffusion}. We collect this in-scene demonstration at 10 Hz.

\subsubsection{Processing and Object Tracking}
\method{} uses object point clouds as observations during policy learning. This representation makes the observations invariant to background changes and visual differences between humans and robots. To obtain these object point clouds, we leverage off-the-shelf computer vision models. For each demonstration, we first segment the objects of interactions in the initial frame using a language prompt with Grounded-SAM~\cite{grounded-sam}. The language prompts used for each task are described in the Appendix~\ref{sec:appendix:tasks}. Next, we track the segmented objects across frames using CoTracker~\cite{co-tracker}, which produces 2D object points for each demonstration. Finally, given per-frame depth, we unproject these 2D points into 3D, effectively obtaining 3D point object point clouds across time.
While this process is straightforward for in-scene demonstrations, the Aria glasses do not provide depth. Therefore, for in-the-wild demonstrations, we use FoundationStereo~\cite{foundation-stereo}, a framework that estimates a disparity map from rectified stereo images and the baseline between the cameras. For this, we use streams from the two front-facing SLAM cameras, rectify them, and use the translation norm provided by the Aria glasses as the baseline $B$. These inputs are passed to FoundationStereo to obtain a disparity map $d$ with respect to the left SLAM camera. Using classical stereo geometry, the depth relative to the left frame, $Z$, is then recovered as:
$$
Z = \frac{f \cdot B}{d},
$$
where $f$ is the focal length of the left camera. 2D object tracks can then be unprojected to this estimated depth for in-the-wild demonstrations. For consistency, we transform all in-scene points into the robot base frame and all in-the-wild demonstrations into the world frame assigned during data collection. This ensures that all points lie on a similar horizontal plane since Aria glasses use the gravity vector to assign the world frame (explained in Section~\ref{sec:method:data_collection}).

\subsubsection{Domain Alignment}
3D object tracks calculated with respect to the Aria glasses vary across demonstrations in the in-the-wild dataset, particularly due to differences in the height of the manipulated objects or the user collecting the data. To address this issue, we transform all 3D points into the robot base frame before training policies, using the in-scene demonstrations as an anchor.

Each demonstration consists of a trajectory of object points $\mathcal{O}^t \in \mathbb{R}^{N\times3}$ and fingertip points $\mathcal{F}^t \in \mathbb{R}^{5\times3}$, where $N$ is the number of objects, fixed at 500 across all tasks. We refer to in-the-wild trajectories as $\mathcal{T}_w = \{\mathcal{O}^t_w,\mathcal{F}^t_w\}$ and in-scene trajectories as $\mathcal{T}_s = \{\mathcal{O}^t_s,\mathcal{F}^t_s\}$.
To transform these trajectories into a uniform space, given a single in-scene trajectory $\mathcal{T}_s$ and an in-the-wild trajectory $\mathcal{T}_w$, we compute the translation between the centroids of the object points in their first frames,
$\Delta \mathcal{O} = \mathcal{O}^0_s - \mathcal{O}^0_w$.
We then translate the in-the-wild trajectory by this offset, yielding
$\hat{\mathcal{T}_w} = \{\mathcal{O}^t + \Delta \mathcal{O}, \mathcal{F}^t + \Delta \mathcal{O}\}$.
This aligns the centroids of the object point clouds. However, since the world frame’s rotation around gravity is assigned arbitrarily, relying solely on this translation can lead to large variations in $z$-axis orientation. This may cause demonstrations where the initial hand pose is fully rotated and object positions appear swapped. Figure illustrating this issue can be found on \website{}.

To estimate a reliable rotation around the $z$-axis, we use the initial hand poses of both trajectories, $\mathcal{F}^0_s$ and $\mathcal{F}^0_w$, and apply the Kabsch algorithm~\cite{kabsch-algorithm} to compute the rigid transform between them. From this transform, we extract the rotation around the $z$-axis $R_z$, and apply it to the in-the-wild demonstrations, yielding the final transformed trajectories:

\begin{align}
    \hat{\mathcal{O}}^t_w &= R_z \cdot \mathcal{O}^t_w + \Delta \mathcal{O} \\
    \hat{\mathcal{F}}^t_w &= R_z \cdot \mathcal{F}^t_w + \Delta \mathcal{O} \\
    \hat{T}_w &= \{ \hat{\mathcal{O}}^t_w,\hat{\mathcal{F}}^t_w \}
\end{align}

We apply this transformation to every in-the-wild demonstration, and both in-the-wild and in-scene demonstrations are then used for policy learning, as described in the next section.  

\subsection{Learning and Deploying Smart Glass Policies on Robots}

\subsubsection{Policy Learning}

To handle visual differences between the robot environment and in the wild human demonstrations, \method{} utilizes transformer-based point-cloud policies and builds on top of the state-of-the-art imitation learning algorithm Point-Policy~\cite{point-policy}. We provide the policy with a trajectory of fingertips $\mathcal{F}^{t-T_o:t}$ and object points $\mathcal{O}^{t - T_o:t}$ as input, and train the model to predict the subsequent fingertip trajectory $\mathcal{F}^{t:t+T_p}$, where $T_o=10$ and $T_p=30$ denote the observation history and prediction horizon, respectively.
In our architecture, the observation history for each point is encoded into a single vector using Vector Neuron Multilayer Perceptrons (MLPs)\cite{deng2021vector}. These differ from regular MLPs in two key ways: (1) points are represented with 3D perceptrons rather than 1D, and (2) they employ SO(3)-equivariant activation layers. We choose vector neuron MLPs due to their demonstrated ability to better capture 3D geometric information\cite{deng2021vector}. 
The flattened vectors are then passed into a transformer encoder as tokens. Positional encoding is learned for only fingertip tokens and not keypoint tokens. The representations output by this encoder are subsequently fed into an MLP to predict the future fingertip trajectory. Mathematically, this can be expressed as follows:

\begin{align}
    \hat{\mathcal{F}}^{t:t+T_p} = \pi \bigl(\mathcal{F}^{t-T_o:t},\mathcal{O}^{t - T_o:t}\bigr).
\end{align}

The entire system is trained end-to-end in a supervised manner using the mean squared error between the predicted and the ground-truth fingertips:

\begin{align}
    \mathcal{L}_{\text{MSE}} = \mathbb{E}\left[ \bigl(\mathcal{F}^{t:t+T_p} - \hat{\mathcal{F}}^{t:t+T_p}\bigr)^2 \right].
\end{align}

In order to improve generalization, we apply augmentations during training. For each datapoint, we uniformly sample a 3D translation in the range $[-30\text{cm}, 30\text{cm}]$, a scaling factor in the range $[0.8, 1.2]$, and a rotation between $[-60^\circ, 60^\circ]$ around the gravity axis. These augmentations are combined into a single transformation, which is then applied consistently to both the input to the model and the ground truth output used to calculate $L_{\text{MSE}}$. Finally, to prevent the model from overfitting to the fingertips, we add Gaussian noise in the range $[-2\text{cm}, 2\text{cm}]$ to the input fingertips, but not to the predicted actions. We train the model for 2000 epochs, which typically takes about 2 hours per task. Our architecture is visualized in Fig.~\ref{fig:method}.\\

\subsubsection{Human Policy $\rightarrow$ Robot Deployment}

\begin{figure}[t]
    \centering
    \includegraphics[width=0.9\linewidth]{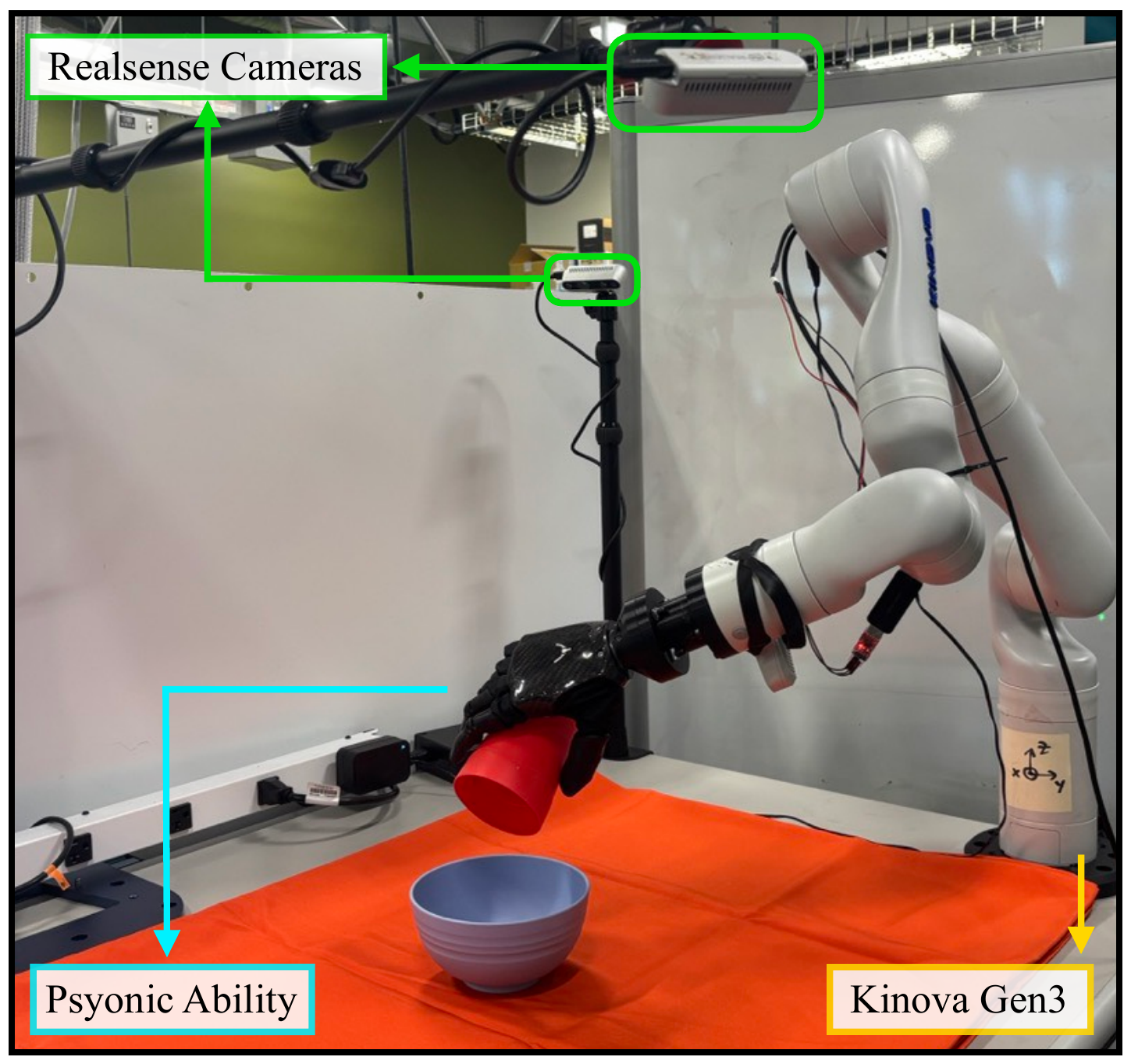}
    \caption{Illustration of our robot setup.}
    \label{fig:robot_setup}
    \vspace{-10pt}
\end{figure}

\begin{figure*}[!t]
    \centering
    \includegraphics[width=\linewidth]{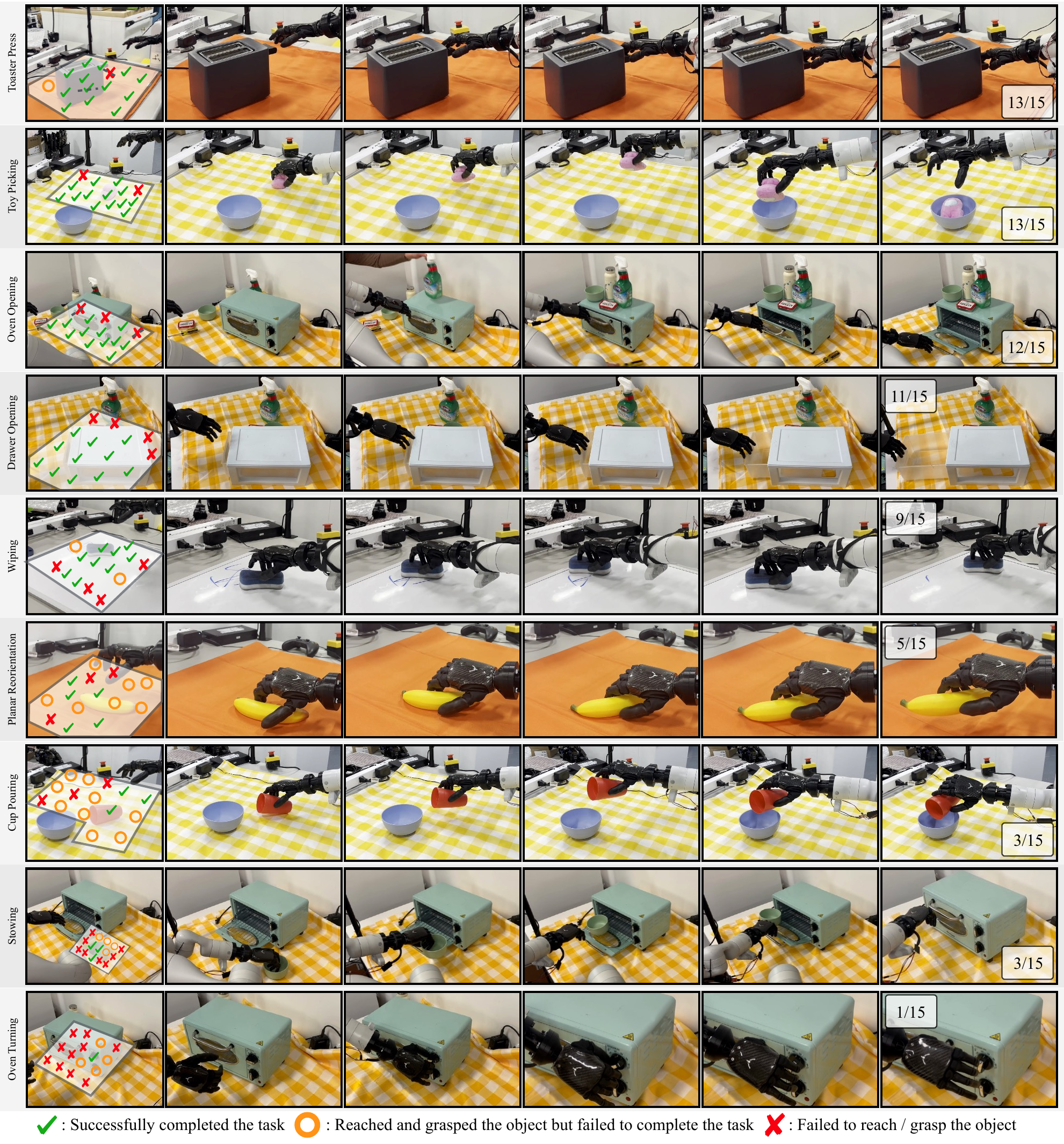}
    \caption{Robot rollouts of \method{} across nine tasks. Spatial generalization is shown in the leftmost column for each task. The meaning of each symbol is explained below the figure. Dotted lines indicate the object's orientation; when not shown, the orientation remains the same as in the showcased rollout. For the Oven Opening task, we showcase \method{}'s performance when there is background disturbance.}
    \label{fig:tasks_rollout}
    \vspace{-12pt}
\end{figure*}

\paragraph{\textbf{Robot Setup}} Our robot setup consists of a single Kinova Gen3 robot arm~\cite{kinova-gen3} with 7 degrees of freedom (DOF) and a Psyonic Ability Hand with five fingers~\cite{ability-hand}. The Ability Hand has six DOFs: one in each finger and two in the thumb. It is designed as a prosthetic hand, making it compact and similar in size to a human hand. To observe the robot’s environment, we use two RealSense RGB-D cameras placed around the operation space. Our robot configuration is illustrated in Fig.~\ref{fig:robot_setup}.

\paragraph{\textbf{Inverse Kinematics}}
The kinematics of human arms and hands differ from those of tabletop manipulators and robot hands, making it non-trivial to replay human trajectories on a robot. Although the Ability Hand’s small size reduces this embodiment gap, the lack of wrist joints in tabletop manipulators means that naively moving the arm and hand separately often leads to infeasible configurations. To address this, we implemented a custom full arm–hand inverse kinematics (IK) module $\mathcal{I}$, similar to \cite{hudor}. Given desired fingertips $\mathcal{F}^{t+1} \in \mathbb{R}^{5\times3}$ and current Kinova and Ability joints $\mathcal{J}^t \in \mathbb{R}^{13}$, the module outputs next joint angles $\mathcal{J}^{t+1} = \mathcal{I}(\mathcal{F}^{t+1}, \mathcal{J}^t)$. The policy predicts fingertips as actions, and the resulting joint angles are applied to the robot during deployment. As in training, we segment and track objects in 3D to obtain object points and use forward kinematics to compute the fingertips.

\paragraph{\textbf{Practical Implementation Details}} Since the human demonstrations do not include force information, for tasks involving grasping, we set a grasping threshold: if the distance between the predicted thumb and any other finger position is less than 5 cm, the fingers are moved closer together. This helps mimic the force that humans apply during grasps.

\section{Experimental Evaluation}
\label{sec:experiments}

\begin{figure*}[]
    \centering
    \includegraphics[width=\linewidth]{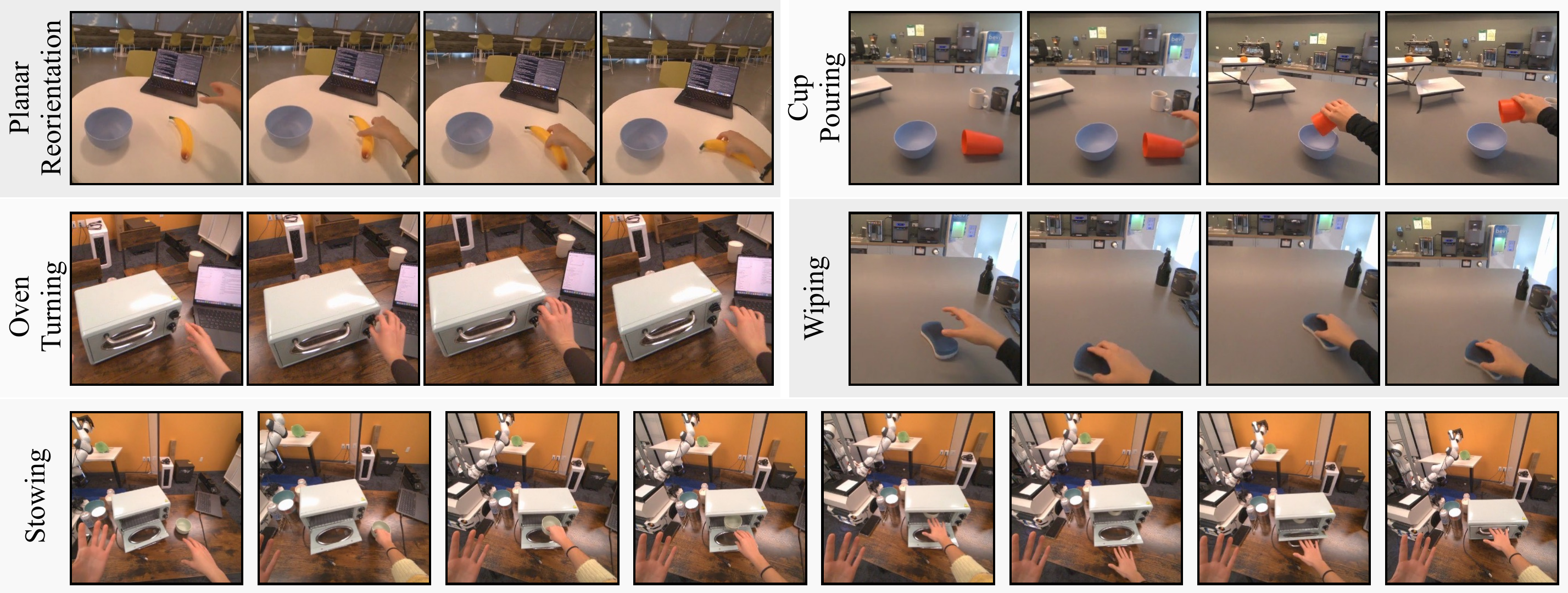}
    \caption{Visualization of in-the-wild human demonstrations collected for different tasks. These are collected with natural human motions and with the right hand performing the respective tasks (no additional sensors on the humans or the environments, except Aria glasses).}
    \label{fig:human_tasks}
    \vspace{-12pt}
\end{figure*}

We perform various real robot experiments and compare \method{} against multiple baselines to answer the following questions:
\begin{enumerate}
    \item How important are the different types of data used in \method{}?
    \item How does \method{} compare to image-based approaches for learning from human data? 
    \item How well does \method{} perform when the height of the operation space changes?
    \item How well does \method{} generalize spatially and across different objects?
\end{enumerate}

\subsection{Task Descriptions}
\label{sec:experiments:tasks}

We evaluate \method{} on nine tasks, each chosen to represent a distinct skill or motion modality (wiping, pick-place, reorientation) and to reflect common daily manipulation activities. Robot rollouts, success rates, and spatial generalization results for each task are shown in Fig.~\ref{fig:tasks_rollout}. Human demonstrations used to train different tasks are shown in Fig.~\ref{fig:human_tasks}. We describe each task in detail in the Appendix ~\ref{sec:appendix:tasks}.

\subsection{How important are the different types of data used in \method{}?} 

\method{} is a new framework for learning robot policies by co-training on in-the-wild and in-scene human video demonstrations. In-scene demonstrations are used both to standardize the input observations and to improve the policy. In this section, we evaluate the importance of this recipe.

We compare \method{} against the following baselines and present the results in Table~\ref{tab:data_comparison}:
\begin{enumerate}
    \item \textbf{In-Scene Only}~\cite{hudor}: A policy trained using only a single in-scene demonstration. Unlike HuDOR~\cite{hudor}, we do not apply any reinforcement learning for this baseline.  
    
    \item \textbf{In-The-Wild Only}~\cite{egozero}: A policy trained solely on in-the-wild demonstrations. These demonstrations are recorded with respect to the initial frame of the RGB camera and then transformed into the robot space by measuring the distance from the left camera to the center of the operation space and shifting the points accordingly. The closest approach to this is EgoZero~\cite{egozero}, but our baseline differs in two key ways: (a) we do not use ArUco markers for data transfer, and (b) we perform closed-loop tracking of all the object points.  
    
    \item \textbf{In-Scene Transform and In-The-Wild}~\cite{zeromimic}: A policy that does not use in-scene data during training, but uses the in-scene demonstration for transforming the in-the-wild demonstrations. This baseline is inspired by ZeroMimic~\cite{zeromimic} that trains policies with in-the-wild human videos and uses a single in-scene goal image to condition the framework.
    \item \textbf{In-Scene Training and In-The-Wild}: A policy that does not use in-scene data for transformation but includes it during training. The transformation is done as described for the \textit{In-The-Wild Only} baseline.
\end{enumerate}

\begin{table}[ht]
\caption{Comparison of success of \method{} to policies trained with different datasets. All methods are evaluated in similar deployment scenarios, with minimum of 10 trials each.}
\centering
\begin{tabular}{@{}ccc@{}}
\toprule
Tasks & Toaster Press & Toy Picking   \\
\midrule
In-Scene Only~\cite{hudor}  & \cellcolor{yellow!30}30\% & \cellcolor{yellow!15}10\% \\
In-The-Wild Only~\cite{egozero}  & \cellcolor{yellow!5}0\% & \cellcolor{yellow!5}0\% \\
In-Scene Transform and In-The-Wild~\cite{zeromimic}  & \cellcolor{yellow!5}0\% & \cellcolor{yellow!15}10\% \\
In-Scene Training and In-The-Wild  & \cellcolor{yellow!60}60\% & \cellcolor{yellow!30}20\% \\ 
In-Scene and In-The-Wild (\method{})  & \cellcolor{yellow!90}\textbf{86}\% & \cellcolor{yellow!90}\textbf{86}\% \\
\bottomrule
\end{tabular}
\label{tab:data_comparison}
\end{table}

From these results, we make the following observations:

\textit{In-the-wild demonstrations improve spatial generalization.} The \textit{In-Scene Only} baseline succeeds when objects are placed close to the demonstrated position, but it fails to generalize beyond that location.

\textit{In-scene demonstrations improve training.} Since deployment is performed using RGB-D cameras rather than Aria glasses, the actions predicted by the \textit{In-The-Wild Only} and \textit{In-Scene Transform} baselines appear highly misaligned, leading to behaviors that look out of distribution. 

\textit{In-scene demonstrations help transform in-the-wild demonstrations.} The in-the-wild data used here is collected on different surfaces, with varying heights and different initial head frames. This makes the transformation in the \textit{In-The-Wild Only} baseline prone to unstable rotations of the object points, resulting in less accurate policies.

\subsection{How does \method{} compare to image-based approaches for learning from human data?}

\begin{figure}[t]
    \centering
    \includegraphics[width=\linewidth]{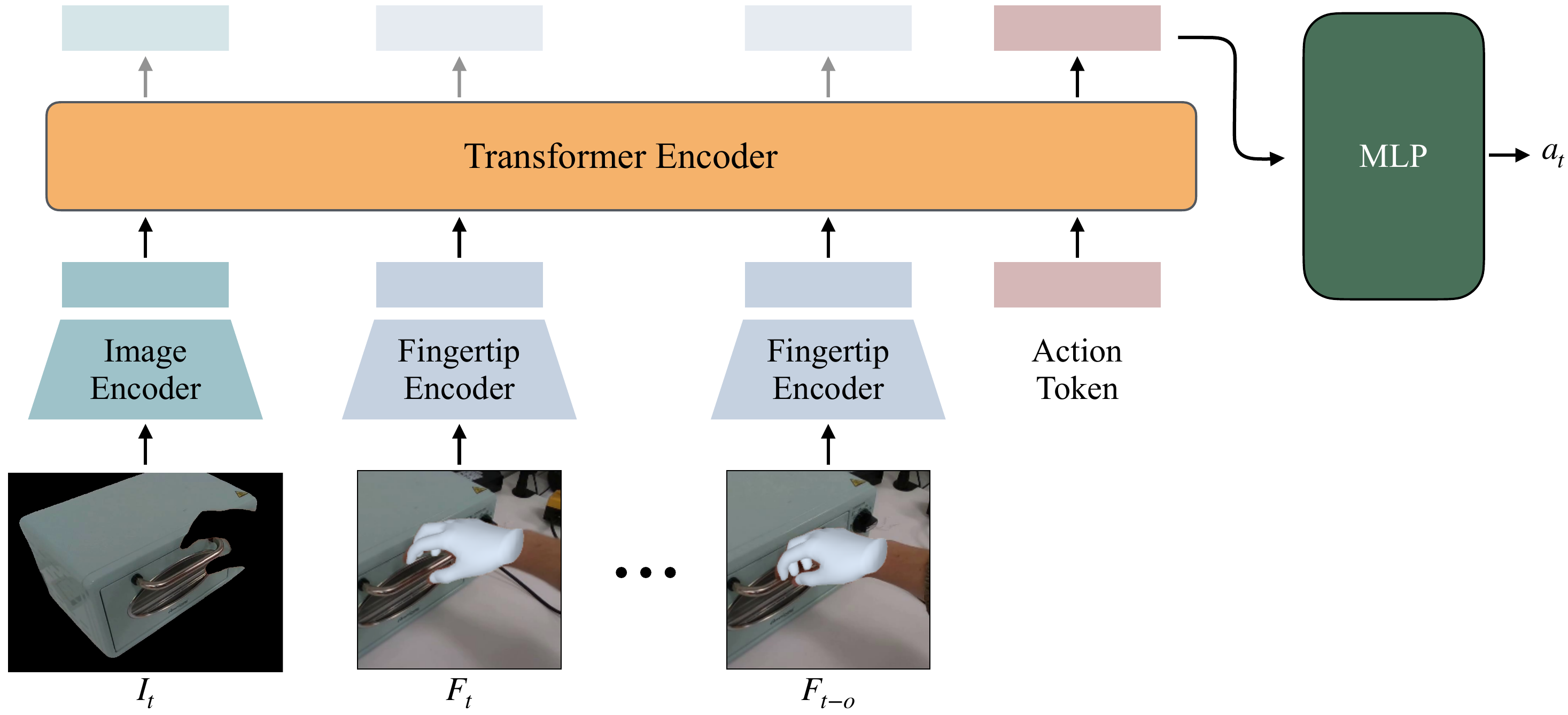}
    \caption{Illustration of BAKU~\cite{haldar2024baku} used in the RGB-based baselines. The fingertip encoders are multilayer perceptrons, and the image encoders are ResNet-18~\cite{resnet} models pretrained on the ImageNet classification task. The action token is set to zeros.}
    \label{fig:image_baseline}
    \vspace{-10pt}
\end{figure}

\method{} uses object-centric point clouds as input to reduce the visual disparity between human and robot observations. Using point clouds and the alignment module in \method{} also improves robustness to viewpoint differences between in-the-wild demonstrations and robot deployment scenarios. To evaluate the impact of using point clouds, we compare \method{} to two image-based architectures on two of our tasks, with the results shown in Table~\ref{tab:image_baseline}. We implement the following baselines:

\begin{enumerate}
    \item \textbf{Masked BAKU:} We segment objects using the same approach as in \method{} and track masks across trajectories using Cutie~\cite{cutie}. We then apply BAKU~\cite{haldar2024baku}, a transformer-based imitation learning architecture, using the masked RGB image of the objects along with the history of fingertip positions. A visualization of this architecture is shown in Fig.~\ref{fig:image_baseline}. In this baseline, we provide fingertip history as input, but only a single RGB frame.
    \item \textbf{Masked BAKU with History:} This version uses the same architecture as Masked BAKU but includes a history of RGB images instead of a single frame.
\end{enumerate}

\begin{table}[ht] \caption{Comparison of success of \method{} to policies trained with RGB images as input. All methods are evaluated in similar deployment scenarios, with 15 trials.} \centering \begin{tabular}{@{}ccc@{}} \toprule Tasks & Oven Opening & Drawer Opening \\ \midrule Masked BAKU & \cellcolor{yellow!30} 6/15 & \cellcolor{yellow!10} 1/15 \\ Masked BAKU with History & \cellcolor{yellow!0} 0/15 & \cellcolor{yellow!0} 0/15 \\ \method{} & \cellcolor{yellow!90}\textbf{12/15} & \cellcolor{yellow!90}\textbf{11/15} \\ \bottomrule \end{tabular} \label{tab:image_baseline} \end{table}

Both baselines are trained on the same dataset as \method{}. We observe that \method{} outperforms these image-based baselines on both tasks. Within the in-the-wild demonstrations, the human head naturally moves, whereas the robot’s camera remains fixed during deployment. This discrepancy causes the Masked BAKU with History inputs to fall out of distribution relative to the training data, causing the policies to perform extremely poorly. Masked BAKU performs better, succeeding in nearly half of the trials; however, we still observe that viewpoint disparity between human demonstrations and robot deployment negatively affects performance. These demonstrate the importance of ingesting 3D inputs, and point tracks instead of images, for effective human-to-robot transfer.

\subsection{How does \method{} perform when the height of the operation space changes?} 

\begin{figure}[t]
    \centering
    \includegraphics[width=\linewidth]{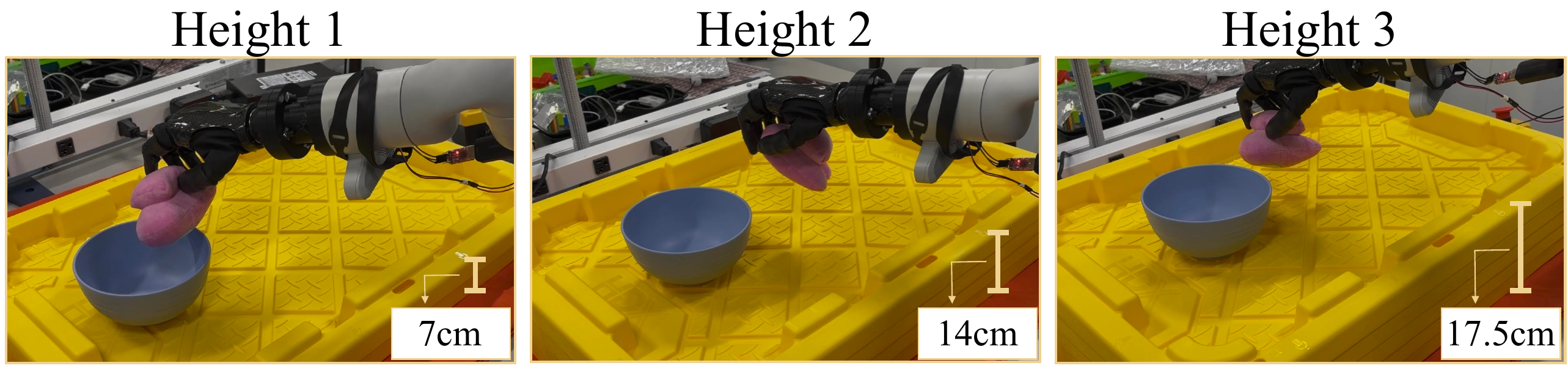}
    \caption{Illustration of the height experiments. Each yellow plate is 3.5 cm tall. \textit{Height 1} consists of 2 plates, \textit{Height 2} of 4 plates, and \textit{Height 3} of 5 plates. Thus, Height 1 is closest to the original deployment scenario, while Height 3 is the furthest.}
    \label{fig:height}
    \vspace{-15pt}
\end{figure}

\method{} does not assume prior knowledge about the height of the manipulated object, the data collector, or the robot’s operation space. To demonstrate its use in operation spaces with different heights, we placed 3.5 cm tall plates on top of the robot’s desk to create three height levels, as illustrated in Fig.~\ref{fig:height}. For each height level, we collect an additional in-scene human demonstration for alignment (requiring less than a minute to collect), as described in Section~\ref{sec:method:data_collection} and use the same human data originally collected in-the-wild. We show the results in Table~\ref{tab:height}.

\begin{table}[ht]
\caption{Success rate of \method{} deployed on plates with different height levels.}
\centering
\begin{tabular}{@{}ccc@{}}
\toprule
Tasks & Toy Picking & Wiping   \\
\midrule
Height 1 & 5/10 & 5/10 \\
Height 2 &  6/10 & 5/10 \\
Height 3 &  2/10 & 8/10 \\ \bottomrule
\end{tabular}
\label{tab:height}
\vspace{-10pt}
\end{table}

We find that the resulting policies perform robustly across tasks, reliably generalizing across heights. This demonstrates the flexibility of \method{} in transferring manipulation skills from in-the-wild data to new scenarios with minimal human effort. Occasional failures arise when an in-scene human demonstration trajectory diverges significantly from the distribution of in-the-wild data. For example, in the Toy Picking task at Height 3, the in-scene demonstration brought the toy unusually close to the bowl. This atypical trajectory led the policy to reproduce the behavior during deployment, causing the toy to push the bowl.

\subsection{How does \method{} generalize to different objects?}

\begin{figure}[t]
    \centering
    \includegraphics[width=\linewidth]{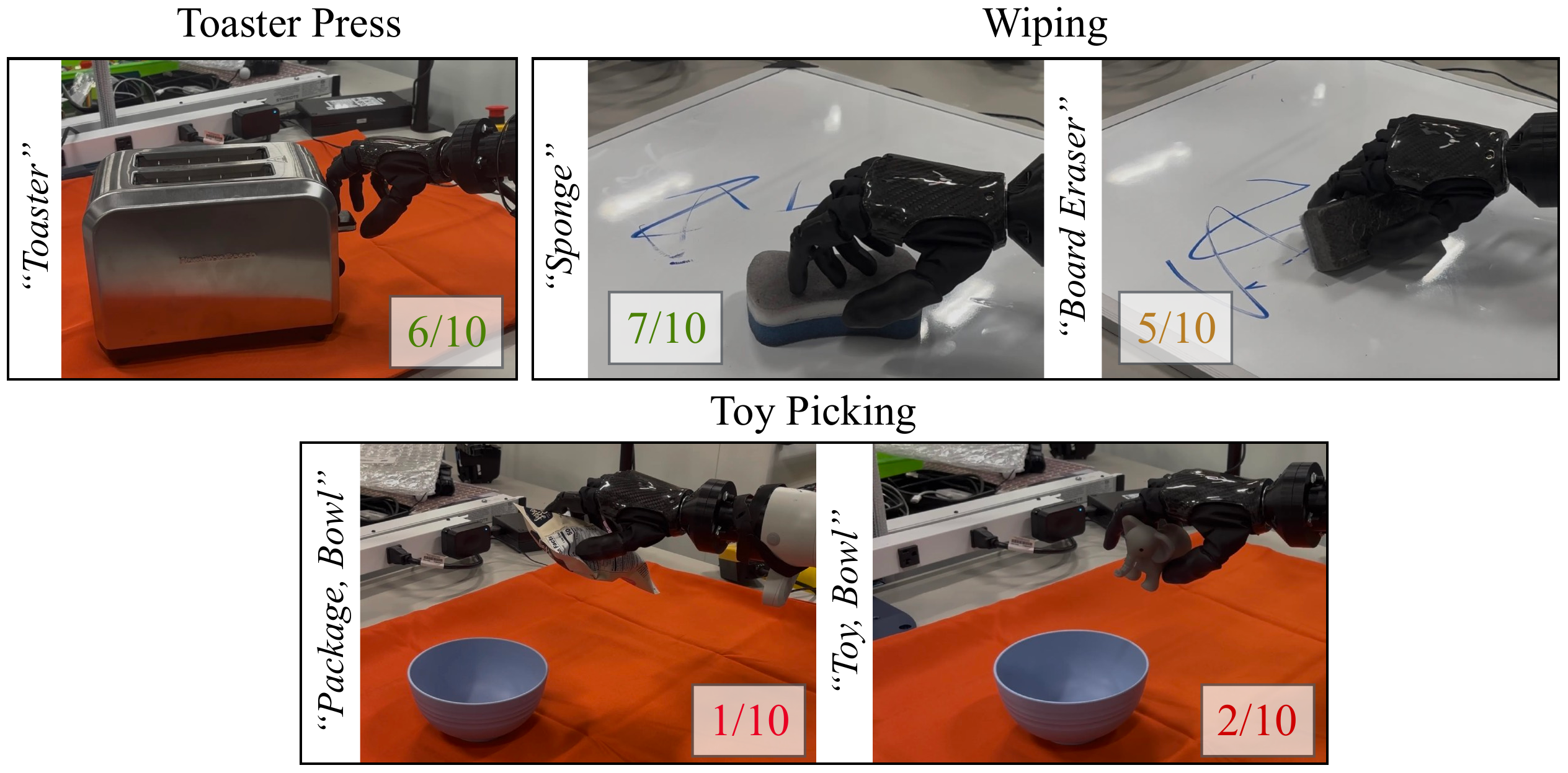}
    \caption{Generalization experiments on Toy Picking, Toaster Press and Wiping tasks. Language prompts used to track the objects are showcased next to each object.}
    \label{fig:generalization}
    \vspace{-10pt}
\end{figure}

We evaluate the generalization of \method{} by testing policies on novel objects across three tasks. Here, we do not train any new policies; instead, we deploy existing ones zero-shot in environments with new objects while prompting GroundedSAM with new task keywords. The success rates and corresponding text prompts are shown in Fig.~\ref{fig:generalization}. We observe that for objects with similar shapes, such as the new toaster or the white eraser, \method{} generalizes well. However, when the shape and weight of the objects differ significantly—such as a popcorn package compared to the toy or a board eraser compared to the sponge—\method{} struggles to generalize.

\section{Discussion, Limitations, Conclusion}

In this work, we presented \method{}, a new framework that leverages capabilities of Aria Gen 2 glasses to learn point-based multi-fingered policies from explicitly in-the-wild human demonstrations. 

While promising, we observe three limitations. First, our framework cannot easily integrate force feedback, since hand pose estimation alone cannot capture this information, which is often crucial for accurate dexterous manipulation~\cite{sparsh-x,sharma2025sparshskin,tdex}. This could be addressed by integrating other wearables, such as EMG sensors or force-estimating gloves. Second, the Aria Gen 2 glasses exhibit a slight difference in shutter timing between the RGB and SLAM cameras. Rapid head movements during data collection can therefore cause misalignment between the object’s pixels in the RGB image and the corresponding depth in SLAM. To mitigate this, we currently instruct data collectors to avoid rapid head movements, though alternative solutions include using more robust 3D object tracking algorithms~\cite{stereo4d} or fitting and tracking a mesh representation of the object~\cite{foundationpose}. Finally, during deployment we currently use Realsense cameras, which causes the keypoints collected with Aria glasses to differ slightly from those observed at deployment. The reason we are not yet streaming Aria input is the difficulty of obtaining real-time depth estimates with FoundationStereo. However, this is an ongoing effort and we believe that with sufficient optimizations, we can receive near real-time depth.
\section*{Acknowledgements}
We thank our amazing colleagues at Meta FAIR and Reality Labs for helpful discussions.

{\scriptsize
\bibliographystyle{IEEEtranN}
\bibliography{ref}

@string{rss = "RSS"}

@string{corl = "CoRL"}

@string{icra = "ICRA"}

@string{cvpr = "CVPR"}

@string{eccv = "ECCV"}

@string{tog = "Transcations on Graphics"}

@string{neurips = "NeurIPS"}

@misc{resnet,
      title={Deep Residual Learning for Image Recognition}, 
      author={Kaiming He and Xiangyu Zhang and Shaoqing Ren and Jian Sun},
      year={2015},
      eprint={1512.03385},
      archivePrefix={arXiv},
      primaryClass={cs.CV},
      url={https://arxiv.org/abs/1512.03385}, 
}

@misc{cutie,
      title={Putting the Object Back into Video Object Segmentation}, 
      author={Ho Kei Cheng and Seoung Wug Oh and Brian Price and Joon-Young Lee and Alexander Schwing},
      year={2024},
      eprint={2310.12982},
      archivePrefix={arXiv},
      primaryClass={cs.CV},
      url={https://arxiv.org/abs/2310.12982}, 
}

@article{haldar2024baku,
                title={BAKU: An Efficient Transformer for Multi-Task Policy Learning},
                author={Haldar, Siddhant and Peng, Zhuoran and Pinto, Lerrel},
                journal={arXiv preprint arXiv:2406.07539},
                year={2024}
              }

@inproceedings{hamer,
  title={Reconstructing hands in 3d with transformers},
  author={Pavlakos, Georgios and Shan, Dandan and Radosavovic, Ilija and Kanazawa, Angjoo and Fouhey, David and Malik, Jitendra},
  booktitle=cvpr,
  year={2024}
}

@inproceedings{srirama2024hrp,
  title={HRP: Human Affordances for Robotic Pre-Training},
  author={Srirama, Mohan Kumar and Dasari, Sudeep and Bahl, Shikhar and Gupta, Abhinav},
  booktitle=rss,
  year={2024}
}

@article{point-policy,
    title={Point Policy: Unifying Observations and Actions with Key Points for Robot Manipulation},
    author={Haldar, Siddhant and Pinto, Lerrel},
    journal={arXiv preprint arXiv:2502.20391},
    year={2025}
  }

@misc{demo-diffusion,
      title={DemoDiffusion: One-Shot Human Imitation using pre-trained Diffusion Policy},
      author={Sungjae Park and Homanga Bharadhwaj and Shubham Tulsiani},
      year={2025},
      eprint={2506.20668},
      archivePrefix={arXiv},
      primaryClass={cs.RO},
      url={https://arxiv.org/abs/2506.20668},
}

@misc{kinova-gen3,
    year = {2010},
    key = "kinova",
    title = "Kinova Gen 3",
    url = {https://www.kinovarobotics.com/product/gen3-robots}

}

@misc{ability-hand,
    year = {2023},
    key = "ability-hand",
    title = "Psyonic Ability Hand",
    url = {https://www.psyonic.io/ability-hand}

}

@misc{egozero,
      title={EgoZero: Robot Learning from Smart Glasses}, 
      author={Vincent Liu and Ademi Adeniji and Haotian Zhan and Raunaq Bhirangi and Pieter Abbeel and Lerrel Pinto},
      year={2025},
      eprint={2505.20290},
      archivePrefix={arXiv},
      primaryClass={cs.RO},
      url={https://arxiv.org/abs/2505.20290}, 
}

@misc{aria-gen-2,
    year = {2025},
    key = "aria-gen-2",
    title = "Aria Gen 2 Glasses",
    url = {https://ai.meta.com/blog/aria-gen-2-research-glasses-under-the-hood-reality-labs/}
}

@article{aria-gen-1,
  title={Project aria: A new tool for egocentric multi-modal ai research},
  author={Engel, Jakob and Somasundaram, Kiran and Goesele, Michael and Sun, Albert and Gamino, Alexander and Turner, Andrew and Talattof, Arjang and Yuan, Arnie and Souti, Bilal and Meredith, Brighid and others},
  journal={arXiv preprint arXiv:2308.13561},
  year={2023}
}

@article{nimble,
  title={Nimble: a non-rigid hand model with bones and muscles},
  author={Li, Yuwei and Zhang, Longwen and Qiu, Zesong and Jiang, Yingwenqi and Li, Nianyi and Ma, Yuexin and Zhang, Yuyao and Xu, Lan and Yu, Jingyi},
  journal={ACM Transactions on Graphics (TOG)},
  volume={41},
  number={4},
  pages={1--16},
  year={2022},
  publisher={ACM New York, NY, USA}
}

@article{mano,
  title={Embodied hands: Modeling and capturing hands and bodies together},
  author={Romero, Javier and Tzionas, Dimitrios and Black, Michael J},
  journal={arXiv preprint arXiv:2201.02610},
  year={2022}
}

@article{kabsch-algorithm,
  title={A purely algebraic justification of the Kabsch-Umeyama algorithm},
  author={Lawrence, Jim and Bernal, Javier and Witzgall, Christoph},
  journal={Journal of research of the National Institute of Standards and Technology},
  volume={124},
  pages={1},
  year={2019}
}

@inproceedings{foundation-stereo,
  title={Foundationstereo: Zero-shot stereo matching},
  author={Wen, Bowen and Trepte, Matthew and Aribido, Joseph and Kautz, Jan and Gallo, Orazio and Birchfield, Stan},
  booktitle={Proceedings of the Computer Vision and Pattern Recognition Conference},
  pages={5249--5260},
  year={2025}
}

@article{co-tracker,
  title={Cotracker3: Simpler and better point tracking by pseudo-labelling real videos},
  author={Karaev, Nikita and Makarov, Iurii and Wang, Jianyuan and Neverova, Natalia and Vedaldi, Andrea and Rupprecht, Christian},
  journal={arXiv preprint arXiv:2410.11831},
  year={2024}
}

@article{grounded-sam,
  title={Grounded sam: Assembling open-world models for diverse visual tasks},
  author={Ren, Tianhe and Liu, Shilong and Zeng, Ailing and Lin, Jing and Li, Kunchang and Cao, He and Chen, Jiayu and Huang, Xinyu and Chen, Yukang and Yan, Feng and others},
  journal={arXiv preprint arXiv:2401.14159},
  year={2024}
}

@article{stereo4d,
  title={Stereo4D: Learning How Things Move in 3D from Internet Stereo Videos}, 
  author={Jin, Linyi and Tucker, Richard and Li, Zhengqi and Fouhey, David and Snavely, Noah and Holynski, Aleksander},
  journal={arXiv preprint},
  year={2024},
}

@inproceedings{hudor,
  title={Bridging the human to robot dexterity gap through object-oriented rewards},
  author={Guzey, Irmak and Dai, Yinlong and Savva, Georgy and Bhirangi, Raunaq and Pinto, Lerrel},
  booktitle={2025 IEEE International Conference on Robotics and Automation (ICRA)},
  pages={3344--3351},
  year={2025},
  organization={IEEE}
}

@misc{egomimic,
  title={EgoMimic: Scaling Imitation Learning via Egocentric Video}, 
  author={Simar Kareer and Dhruv Patel and Ryan Punamiya and Pranay Mathur and Shuo Cheng and Chen Wang and Judy Hoffman and Danfei Xu},
  year={2024},
  eprint={2410.24221},
  archivePrefix={arXiv},
  primaryClass={cs.RO},
  url={https://arxiv.org/abs/2410.24221},
}

@inproceedings{zeromimic,
      title={ZeroMimic: Distilling Robotic Manipulation Skills from Web Videos}, 
      author={Junyao Shi and Zhuolun Zhao and Tianyou Wang and Ian Pedroza and Amy Luo and Jie Wang and Jason Ma and Dinesh Jayaraman},
      year={2025},
      booktitle={International Conference on Robotics and Automation (ICRA)},
}

@article{dexwild,
      title={DexWild: Dexterous Human Interactions for In-the-Wild Robot Policies},
      author={Tao, Tony and Srirama, Mohan Kumar and Liu, Jason Jingzhou and Shaw, Kenneth and Pathak, Deepak},
      journal={Robotics: Science and Systems (RSS)},
      year={2025}}

@article{tdex,
  title={Dexterity from touch: Self-supervised pre-training of tactile representations with robotic play},
  author={Guzey, Irmak and Evans, Ben and Chintala, Soumith and Pinto, Lerrel},
  journal={arXiv preprint arXiv:2303.12076},
  year={2023}
}

@inproceedings{tavi,
  title={See to touch: Learning tactile dexterity through visual incentives},
  author={Guzey, Irmak and Dai, Yinlong and Evans, Ben and Chintala, Soumith and Pinto, Lerrel},
  booktitle={2024 IEEE International Conference on Robotics and Automation (ICRA)},
  pages={13825--13832},
  year={2024},
  organization={IEEE}
}

@inproceedings{foundationpose,
  title={Foundationpose: Unified 6d pose estimation and tracking of novel objects},
  author={Wen, Bowen and Yang, Wei and Kautz, Jan and Birchfield, Stan},
  booktitle={Proceedings of the IEEE/CVF Conference on Computer Vision and Pattern Recognition},
  pages={17868--17879},
  year={2024}
}

@article{sparsh-x,
  title={Tactile Beyond Pixels: Multisensory Touch Representations for Robot Manipulation},
  author={Higuera, Carolina and Sharma, Akash and Fan, Taosha and Bodduluri, Chaithanya Krishna and Boots, Byron and Kaess, Michael and Lambeta, Mike and Wu, Tingfan and Liu, Zixi and Hogan, Francois Robert and others},
  journal={arXiv preprint arXiv:2506.14754},
  year={2025}
}

@article{sharma2025sparshskin,
  title={Self-supervised perception for tactile skin covered dexterous hands},
  author={Sharma, Akash and Higuera, Carolina and Bodduluri, Chaithanya Krishna and Liu, Zixi and Fan, Taosha and Hellebrekers, Tess and Lambeta, Mike and Boots, Byron and Kaess, Michael and Wu, Tingfan and others},
  journal={arXiv preprint arXiv:2505.11420},
  year={2025}
}

@misc{wu2023unleashing,
      title={Unleashing Large-Scale Video Generative Pre-training for Visual Robot Manipulation}, 
      author={Hongtao Wu and Ya Jing and Chilam Cheang and Guangzeng Chen and Jiafeng Xu and Xinghang Li and Minghuan Liu and Hang Li and Tao Kong},
      year={2023},
      eprint={2312.13139},
      archivePrefix={arXiv},
      primaryClass={cs.RO}}

@article{banerjee2024hot3d,
  title={{HOT3D}: Hand and Object Tracking in {3D} from Egocentric Multi-View Videos},
  author={Banerjee, Prithviraj and Shkodrani, Sindi and Moulon, Pierre and Hampali, Shreyas and Han, Shangchen and Zhang, Fan and Zhang, Linguang and Fountain, Jade and Miller, Edward and Basol, Selen and Newcombe, Richard and Wang, Robert and Engel, Jakob Julian and Hodan, Tomas},
  journal={CVPR},
  year={2025}
}

@article{zorin2025ruka,
  title={RUKA: Rethinking the Design of Humanoid Hands with Learning},
  author={Zorin, Anya and Guzey, Irmak and Yan, Billy and Iyer, Aadhithya and Kondrich, Lisa and Bhattasali, Nikhil X. and Pinto, Lerrel},
  journal={Robotics: Science and Systems (RSS)},
  year={2025}
}

@article{shaw2023leap,
  title={Leap hand: Low-cost, efficient, and anthropomorphic hand for robot learning},
  author={Shaw, Kenneth and Agarwal, Ananye and Pathak, Deepak},
  journal={arXiv preprint arXiv:2309.06440},
  year={2023}
}

@article{akkaya2019solving,
  title={Solving rubik's cube with a robot hand},
  author={Akkaya, Ilge and Andrychowicz, Marcin and Chociej, Maciek and Litwin, Mateusz and McGrew, Bob and Petron, Arthur and Paino, Alex and Plappert, Matthias and Powell, Glenn and Ribas, Raphael and others},
  journal={arXiv preprint arXiv:1910.07113},
  year={2019}
}

@inproceedings{shridhar2023perceiver,
  title={Perceiver-actor: A multi-task transformer for robotic manipulation},
  author={Shridhar, Mohit and Manuelli, Lucas and Fox, Dieter},
  booktitle={Conference on Robot Learning},
  pages={785--799},
  year={2023},
  organization={PMLR}
}

@article{gervet2023act3d,
  title={Act3d: 3d feature field transformers for multi-task robotic manipulation},
  author={Gervet, Theophile and Xian, Zhou and Gkanatsios, Nikolaos and Fragkiadaki, Katerina},
  journal={arXiv preprint arXiv:2306.17817},
  year={2023}
}

@inproceedings{ze2023gnfactor,
  title={Gnfactor: Multi-task real robot learning with generalizable neural feature fields},
  author={Ze, Yanjie and Yan, Ge and Wu, Yueh-Hua and Macaluso, Annabella and Ge, Yuying and Ye, Jianglong and Hansen, Nicklas and Li, Li Erran and Wang, Xiaolong},
  booktitle={Conference on robot learning},
  pages={284--301},
  year={2023},
  organization={PMLR}
}

@article{chen2024arcap,
            title={ARCap: Collecting High-quality Human Demonstrations for Robot Learning with Augmented Reality Feedback},
            author={Chen, Sirui and Wang, Chen and Nguyen, Kaden and Fei-Fei, Li and Liu, C Karen},
            journal={arXiv preprint arXiv:2410.08464},
            year={2024}
          }

@inproceedings{arunachalam2023holo,
  title={Holo-dex: Teaching dexterity with immersive mixed reality},
  author={Arunachalam, Sridhar Pandian and G{\"u}zey, Irmak and Chintala, Soumith and Pinto, Lerrel},
  booktitle=icra,
  year={2023},

}

@inproceedings{yang2024ace,
  title={ACE: A Cross-Platform Visual-Exoskeletons System for Low-Cost Dexterous Teleoperation},
  author={Yang, Shiqi and Liu, Minghuan and Qin, Yuzhe and Ding, Runyu and Li, Jialong and Cheng, Xuxin and Yang, Ruihan and Yi, Sha and Wang, Xiaolong},
  booktitle=corl,
  year={2024}
}

@inproceedings{pavlakos2024reconstructing,
  title={Reconstructing hands in 3d with transformers},
  author={Pavlakos, Georgios and Shan, Dandan and Radosavovic, Ilija and Kanazawa, Angjoo and Fouhey, David and Malik, Jitendra},
  booktitle=cvpr,
  year={2024}
}

@inproceedings{shaw2023videodex,
  title={Videodex: Learning dexterity from internet videos},
  author={Shaw, Kenneth and Bahl, Shikhar and Pathak, Deepak},
  booktitle=corl,
  year={2023},
}

@inproceedings{wang2024dexcap,
  title={Dexcap: Scalable and portable mocap data collection system for dexterous manipulation},
  author={Wang, Chen and Shi, Haochen and Wang, Weizhuo and Zhang, Ruohan and Fei-Fei, Li and Liu, C Karen},
  booktitle=rss,
  year={2024}
}

@inproceedings{iyer2024open,
  title={Open teach: A versatile teleoperation system for robotic manipulation},
  author={Iyer, Aadhithya and Peng, Zhuoran and Dai, Yinlong and Guzey, Irmak and Haldar, Siddhant and Chintala, Soumith and Pinto, Lerrel},
  booktitle=corl,
  year={2024},
}

@inproceedings{vrb,
  title={Affordances from human videos as a versatile representation for robotics},
  author={Bahl, Shikhar and Mendonca, Russell and Chen, Lili and Jain, Unnat and Pathak, Deepak},
  booktitle=cvpr,
  year={2023}
}

@inproceedings{track2act,
  title={Track2act: Predicting point tracks from internet videos enables generalizable robot manipulation},
  author={Bharadhwaj, Homanga and Mottaghi, Roozbeh and Gupta, Abhinav and Tulsiani, Shubham},
  booktitle=eccv,
  year={2024},
}

@inproceedings{r3m,
  title={R3m: A universal visual representation for robot manipulation},
  author={Nair, Suraj and Rajeswaran, Aravind and Kumar, Vikash and Finn, Chelsea and Gupta, Abhinav},
  booktitle=corl,
  year={2022}
}

@article{pvr2,
  title={Masked visual pre-training for motor control},
  author={Xiao, Tete and Radosavovic, Ilija and Darrell, Trevor and Malik, Jitendra},
  journal={arXiv preprint arXiv:2203.06173},
  year={2022}
}

@article{wang2023mimicplay,
  title={MimicPlay: Long-Horizon Imitation Learning by Watching Human Play},
  author={Wang, Chen and Fan, Linxi and Sun, Jiankai and Zhang, Ruohan and Fei-Fei, Li and Xu, Danfei and Zhu, Yuke and Anandkumar, Anima},
  journal={arXiv preprint arXiv:2302.12422},
  year={2023}
}

@inproceedings{rt1,
  title={Rt-1: Robotics transformer for real-world control at scale},
  author={Brohan, Anthony and Brown, Noah and Carbajal, Justice and Chebotar, Yevgen and Dabis, Joseph and Finn, Chelsea and Gopalakrishnan, Keerthana and Hausman, Karol and Herzog, Alex and Hsu, Jasmine and others},
 booktitle=rss,  
year={2023}
}

@article{cacti,
  title={CACTI: A Framework for Scalable Multi-Task Multi-Scene Visual Imitation Learning},
  author={Mandi, Zhao and Bharadhwaj, Homanga and Moens, Vincent and Song, Shuran and Rajeswaran, Aravind and Kumar, Vikash},
  journal={arXiv preprint arXiv:2212.05711},
  year={2022}
}

@article{chen2025web2grasp,
  title={Web2Grasp: Learning Functional Grasps from Web Images of Hand-Object Interactions},
  author={Chen, Hongyi and Yao, Yunchao and Ye, Yufei and Xu, Zhixuan and Bharadhwaj, Homanga and Wang, Jiashun and Tulsiani, Shubham and Erickson, Zackory and Ichnowski, Jeffrey},
  journal={arXiv preprint arXiv:2505.05517},
  year={2025}
}

@inproceedings{ego4d,
  title={Ego4d: Around the world in 3,000 hours of egocentric video},
  author={Grauman, Kristen and Westbury, Andrew and Byrne, Eugene and Chavis, Zachary and Furnari, Antonino and Girdhar, Rohit and Hamburger, Jackson and Jiang, Hao and Liu, Miao and Liu, Xingyu and others},
  booktitle=cvpr,
  year={2022}
}

@inproceedings{he2022masked,
  title={Masked autoencoders are scalable vision learners},
  author={He, Kaiming and Chen, Xinlei and Xie, Saining and Li, Yanghao and Doll{\'a}r, Piotr and Girshick, Ross},
  booktitle=cvpr,
  year={2022}
}

@inproceedings{ye2022s,
  title={What's in your hands? 3D Reconstruction of Generic Objects in Hands},
  author={Ye, Yufei and Gupta, Abhinav and Tulsiani, Shubham},
  booktitle=cvpr,
  year={2022}
}

@inproceedings{cotracker,
  title={Cotracker: It is better to track together},
  author={Karaev, Nikita and Rocco, Ignacio and Graham, Benjamin and Neverova, Natalia and Vedaldi, Andrea and Rupprecht, Christian},
  booktitle=eccv,
  year={2024}
}

@inproceedings{tapir,
  title={Tap-vid: A benchmark for tracking any point in a video},
  author={Doersch, Carl and Gupta, Ankush and Markeeva, Larisa and Recasens, Adri{\`a} and Smaira, Lucas and Aytar, Yusuf and Carreira, Jo{\~a}o and Zisserman, Andrew and Yang, Yi},
  booktitle=neurips,
  year={2022}
}

@inproceedings{voltron,
  title={Language-driven representation learning for robotics},
  author={Karamcheti, Siddharth and Nair, Suraj and Chen, Annie S and Kollar, Thomas and Finn, Chelsea and Sadigh, Dorsa and Liang, Percy},
  booktitle=rss,
  year={2023}
}

@inproceedings{grauman2024ego,
  title={Ego-exo4d: Understanding skilled human activity from first-and third-person perspectives},
  author={Grauman, Kristen and Westbury, Andrew and Torresani, Lorenzo and Kitani, Kris and Malik, Jitendra and Afouras, Triantafyllos and Ashutosh, Kumar and Baiyya, Vijay and Bansal, Siddhant and Boote, Bikram and others},
  booktitle=cvpr,
  year={2024}
}

@inproceedings{lepert2024shadow,
        title={Shadow: Leveraging Segmentation Masks for Zero-Shot Cross-Embodiment Policy Transfer},
        author={Marion Lepert and Ria Doshi and Jeannette Bohg},
        booktitle = {Conference on Robot Learning (CoRL)},
        address  = {Munich, Germany},
        year = {2024},
  }

@misc{bharadhwaj2023generalizable,
      title={Towards Generalizable Zero-Shot Manipulation via Translating Human Interaction Plans}, 
      author={Homanga Bharadhwaj and Abhinav Gupta and Vikash Kumar and Shubham Tulsiani},
      year={2023},
      eprint={2312.00775},
      archivePrefix={arXiv},
      primaryClass={cs.RO}
}

@inproceedings{deng2021vector,
  title={Vector neurons: A general framework for so (3)-equivariant networks},
  author={Deng, Congyue and Litany, Or and Duan, Yueqi and Poulenard, Adrien and Tagliasacchi, Andrea and Guibas, Leonidas J},
  booktitle={Proceedings of the IEEE/CVF International Conference on Computer Vision},
  pages={12200--12209},
  year={2021}
}
}





\newpage
\appendix
\label{sec:appendix}

\subsection{Task Descriptions}
\label{sec:appendix:tasks}

In this section, we describe each task in detail.

\paragraph{\textbf{Toaster Press}} The robot must locate and push down the lever of a bread toaster. The toaster is positioned within a $30,\text{cm} \times 50,\text{cm}$ area. The text prompt used is \textit{toaster}.

\paragraph{\textbf{Toy Picking}} The robot must locate and pick up a soft pink toy, then drop it into a bowl. The toy is positioned within a $30,\text{cm} \times 30,\text{cm}$ area, while the bowl remains fixed. The text prompts used are \textit{bowl} and \textit{pink toy}.

\paragraph{\textbf{Oven Opening}} The robot must locate a toaster oven and open its door by pulling its lever. The oven is positioned within a $50,\text{cm} \times 30,\text{cm}$ area. The text prompt used is \textit{toaster oven}.

\paragraph{\textbf{Drawer Opening}} The robot must locate a white storage drawer and slide it open. The drawer is positioned within a $50,\text{cm} \times 30,\text{cm}$ area. The text prompt used is \textit{white box}.

\paragraph{\textbf{Wiping}} The robot must locate a sponge and wipe the board. The sponge is positioned within a $30,\text{cm} \times 30,\text{cm}$ area. The demonstrations do not specify where to wipe; wiping motions are collected arbitrarily. Success is therefore defined by whether the robot achieves a stable grasp of the sponge and wipes some portion of the board. The text prompt used is \textit{sponge}.

\paragraph{\textbf{Planar Reorientation}} The robot must locate a banana, reorient it in place, and pick it up. The banana is positioned within a $30,\text{cm} \times 30,\text{cm}$ area. The text prompt used is \textit{banana}.

\paragraph{\textbf{Cup Pouring}} The robot must locate a red cup, pick it up, and pour its contents into a bowl. The cup is positioned within a $30,\text{cm} \times 30,\text{cm}$ area, while the bowl remains fixed. The text prompts used are \textit{red cup} and \textit{bowl}.

\paragraph{\textbf{Stowing}} The robot must locate a bowl, pick it up, place it inside a toaster oven, and close the oven door. This is a long-horizon task involving multiple skills: picking up a rigid bowl, placing it in a spatially constrained location, and closing the oven door. The bowl is positioned within a $20,\text{cm} \times 20,\text{cm}$ area, while the oven remains fixed. The text prompts used are \textit{toaster oven} and \textit{bowl}.

\paragraph{\textbf{Knob Rotating}} The robot must locate the temperature knob of a toaster oven and rotate it 90 degrees. The toaster oven is positioned within a $20,\text{cm} \times 20,\text{cm}$ area. The text prompt used is \textit{toaster oven}.

\end{document}